\documentclass[preprint]{elsarticle}
\usepackage[utf8]{inputenc}
\usepackage{float} 
\usepackage{hyperref}
\usepackage{caption}
\usepackage{subcaption}
\usepackage{amssymb}
\usepackage{amsmath}
\usepackage{amsthm}
\usepackage{nccmath}
\usepackage{graphicx}
\usepackage{color}
\usepackage{multirow}
\usepackage{mathtools}
\usepackage{algpseudocode}
\usepackage{algorithm}
\usepackage{booktabs}
\usepackage[table,xcdraw]{xcolor}

\hypersetup{colorlinks=false,pdfborder={0 0 0},
bookmarksopen,bookmarksdepth=3}

\title{Interpreting Deep Learning Models for Epileptic Seizure Detection on EEG signals}
\author[1]{Valentin Gabeff}
\author[1]{Tomas Teijeiro}
\author[1,2]{Marina Zapater}
\author[3]{Leila Cammoun}
\author[4,5]{Sylvain Rheims}
\author[3]{Philippe Ryvlin}
\author[1]{David Atienza}

\address[1]{Embedded Systems Laboratory (ESL), EPFL, Lausanne, Switzerland}
\address[2]{REDS Institute, University of Applied Sciences Western Switzerland (HEIG-VD/HES-SO), Yverdon-les-Bains, Switzerland}
\address[3]{Department of Clinical Neurosciences, Neurology Service, Centre Hospitalier Universitaire Vaudois (CHUV) and University of Lausanne, Lausanne, Switzerland}
\address[4]{Department of Functional Neurology and Epileptology, Hospices Cvils de Lyon and University of Lyon, Lyon, France}
\address[5]{Lyon's Neurosciences Research Center (INSERM U1028/CNRS UMR 5292), Lyon, France}

\date{\today}

\begin{document}


\begin{abstract}
\small
While Deep Learning (DL) is often considered the state-of-the art for Artificial Intel-ligence-based medical decision support, it remains sparsely implemented in clinical practice and poorly trusted by clinicians due to insufficient interpretability of neural network models. We have tackled this issue by developing interpretable DL models in the context of online detection of epileptic seizure, based on EEG signal. This has conditioned the preparation of the input signals, the network architecture, and the post-processing of the output in line with the domain knowledge. Specifically, we focused the discussion on three main aspects: 1) how to aggregate the classification results on signal segments provided by the DL model into a larger time scale, at the seizure-level; 2) what are the relevant frequency patterns learned in the first convolutional layer of different models, and their relation with the delta, theta, alpha, beta and gamma frequency bands on which the visual interpretation of EEG is based; and 3) the identification of the signal waveforms with larger contribution towards the ictal class, according to the activation differences highlighted using the DeepLIFT method. Results show that the kernel size in the first layer determines the interpretability of the extracted features and the sensitivity of the trained models, even though the final performance is very similar after post-processing. Also, we found that amplitude is the main feature leading to an ictal prediction, suggesting that a larger patient population would be required to learn more complex frequency patterns. Still, our methodology was successfully able to generalize patient inter-variability for the majority of the studied population with a classification F1-score of 0.873 and detecting 90\% of the seizures.
\end{abstract}


\maketitle
\section{Introduction}
Epilepsy is a neurological disease characterized by paroxysmal events, called seizures, arising from the abnormal activation of neuronal networks. This abnormal activation translates into changes in the pattern of electrical activity generated by the brain, which can be captured through electroencephalography (EEG). The disease affects 50 million people worldwide, among which 70\% could live seizure-free with appropriate diagnosis and treatment according to the World Health Organization \cite{who_epilepsy_2019}. Conversely, 30\% of patients with epilepsy continue to suffer unpredictably recurring seizures. For these patients, there is a crucial need for the development of devices to detect seizure events \cite{aghaei-lasboo_methods_2016}. A common practice is for the patient to document each seizure in a paper or electronic diary to later provide an appropriate therapy \cite{aghaei-lasboo_methods_2016,fisher_seizure_2012}. Unfortunately, seizure events are often underreported, partly due to seizure-induced amnesia of seizure events \cite{blachut_counting_2015}. This justifies the development of wireless recording devices coupled with an electronic diary to detect, record and eventually predict or forecast seizure events. Existing wearable devices often only detect a fraction of seizures, called generalized tonic-clonic seizures (GTCS), thanks to easy-to-observe GTCS-induced variation of various biosignals, including surface-electromyography (sEMG), electrodermal activity (EDA) and 3D-accelerometry~\cite{conradsen_seizure_2011, ming-zher_poh_continuous_2010,sopic_e-glass_2018}. However, these biosignals do not currently offer a reliable way to detect the majority of non-GTCS seizures~\cite{Ryvlin2020}. EEG is the gold standard method to detect all seizure types in hospitals~\cite{maganti_eeg_2013,roy_deep_2019}, but no reliable wearable EEG is yet available to transfer this approach for very long-term monitoring at home. Yet, innovative wearables are being developed and might allow such monitoring in the near future, stressing the need for EEG-based on-line seizure detection and interpretable DL models.

With the bloom of Deep Learning (DL) in the biomedical field, several methods have been developed to detect and predict seizure events from EEG of epileptic patients primarily recorded during short in-hospital monitoring with standard scalp-EEG or intracerebral electrodes. Some methods used Recurrent Neural Networks (RNN) to account for the temporal nature of EEG~\cite{tsiouris_long_2018}. However, most methods have used Convolutional Neural Networks (CNN) \cite{truong_convolutional_2018, ullah_automated_2018, yuan_multi-view_2019, yuan_novel_2018, de_aguiar_early_2015, antoniades_deep_2016, avcu_seizure_2019}.Not only their architecture and mechanisms have been constantly improved due to their thorough utilization in the field of modern computer vision, but they often provide better results than simple RNN architectures for the classification of EEG time-series~\cite{bai_evaluation_2018}. Though some methods reported excellent performances, most used offline analysis with significant pre-processing and transformation of the EEG signal not compatible with the aim of on-line, long-term, ambulatory low-power operations.

Overcoming the challenge of efficient characterization of seizure events on a large and heterogeneous population of patients is also a crucial step for the transfer to clinical applications. Indeed, current devices suffer from generalization difficulties to unseen patients and they often need to be fine-tuned to each patient as a result of important patient inter-variability in epileptic disorders~\cite{roy_deep_2019}. Training DL networks often requires dividing the input EEG into short segments, typically between 0.5 and 30 seconds~\cite{roy_deep_2019}. Classification metrics of those individual segments is a necessary step to characterize model performance, but if we aim to provide a model for seizure detection, one should also assess the performance on longer stretches of EEG signals carrying transitions between interictal and ictal phases. This is commonly done by processing the short segments of test EEG signals in a time linear fashion and reporting an aggregated performance (e.g. seizure sensitivity). This helps to understand the behavior of the model during continuous monitoring and it is a necessary step for the development of monitoring devices. Most studies report performance on short segments but few additionally report sensitivity, precision and accuracy for complete seizures episodes. We refer to these two levels of performance as follows: ``segment-level'' depicts performance on the short segments of test EEG, while ``seizure-level'' refers to performance on the individual seizure events.

Given that the ``black-box'' nature of DL is a common obstacle to the transfer of applications in clinics \cite{adadi_explainable_2020}, we aim to explore not just the features learned by a DL model but also the input data properties leading to a classification decision with the view to improve the validity of the method and potentially the understanding of the related pathology. Delineating the key characteristics of the input data for an efficient classification can then justify the conception of the model architecture and the choice of processing methods. Recent studies have applied this approach to elucidate the dynamics of DL models on EEG signals~\cite{sturm_interpretable_2016, schirrmeister_deep_2017, hartmann_hierarchical_2018}.

This work is thus a step towards strengthening the relationship between the current knowledge of EEG signals in epileptic disorders and the development of transferable DL methods for characterization of seizure events. In the continuity of the work with the e-Glass as an EEG monitoring wearable device \cite{sopic_e-glass_2018}, our study focuses on electrodes placed over the temporal brain regions and explores performance of the model both at the segment and seizure levels, with an extensive discussion on how these two levels are related. As a result, we report a non-patient-specific online method using raw EEG to detect seizures, and investigate the features learned by the model, providing a visual feedback of the decisive patterns for seizure detection on the EEG signal.

\section{Materials and Methods}

\subsection{Dataset}
The dataset used in this study comes from the REPO$_2$MSE cohort, which characteristics were previously reported~\cite{Rheims2019}. It contains multi-channel scalp-EEG recordings from 568 patients with epilepsy, and annotations of seizure onsets from an experienced epileptologist. 1212 distinct seizures, each in one record file, are available with each record being cut to contain a single seizure with a median of 3.0 minutes of interictal recording before the annotated seizure onset. EEG recordings were either sampled at 256Hz (89.6\%), 512Hz (10.2\%) or 1024 Hz (0.2\%). The median number of files containing a seizure was of 2 per patient, a maximum of 9 and a minimum of 1. EEG recordings were saved in TRC format and later converted using MATLAB to .mat files for subsequent processing in Python. The large number of subjects in this study will allow us to assess whether the models can successfully generalize to unseen patients.

\subsection{Preparing the input}

Pre-processing steps were minimized to ensure an online method. The input to the model is the raw EEG signal cut in overlapping segments of 5 seconds. Although time-frequency representation of the signal is commonly employed \cite{yuan_multi-view_2019, truong_convolutional_2018}, we believe that a careful design of the model architecture taking into account how experts visually interpret raw EEG signals in time and space may perform as well as more explicit representations. We opted for a long kernel size along the time dimension in the first layer to extract the frequency content of each channel and along the channel dimension to detect potential synchronous activity. Model architecture is further described in Section \ref{section:architect}. This study focuses on the detection of seizures at the following four EEG channels  ``F7-T7'', ``F8-T8'', ``T7-P7'' and ``T8-P8'' in order to reflect future long-term recordings using wearable EEG systems like the e-Glass~\cite{sopic_e-glass_2018}. Each channel is down-sampled at 256Hz if not already recorded at this frequency.

\begin{figure}
    \centering
    \includegraphics[width = \textwidth]{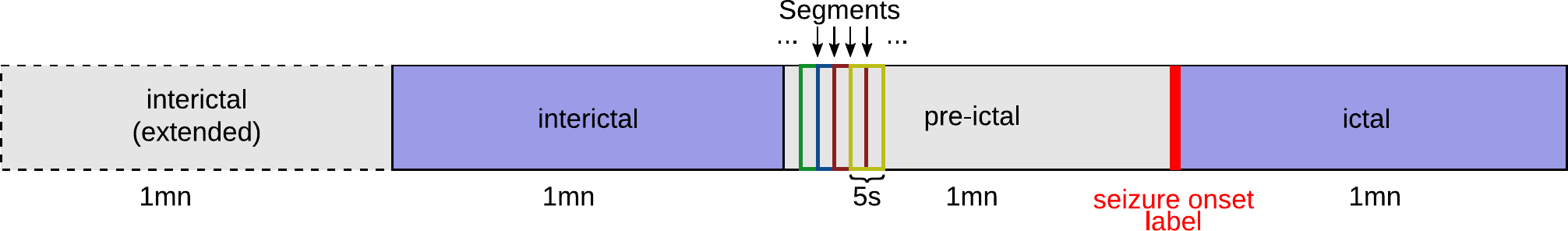}
    \caption{\textbf{Segmentation of EEG signal.} Signals are cut into four parts relative to the seizure onset label. The ictal portion extends up to one minute after seizure onset. The pre-ictal portion is considered as the minute before seizure onset and is preceded by one minute of interictal signal. An additional minute of interictal is added in for post-processing with a difference filter, as explained in Section \ref{section:post_proc}. Ictal and interictal portions (blue) are used for training and computation of metrics. Grey regions are only used for prediction but are not taken into account for the calculation of the evaluation metrics.}
    \label{fig:signal_cutting}
\end{figure}

For each available seizure, we considered one minute of interictal recording and one minute of ictal recording. The interictal segment started two minutes and ended one minute prior to seizure onset to avoid including the immediate pre-ictal minute. The ictal segment started at seizure onset. Some seizures might last less than one minute, leading the ictal segment to include some immediate post-ictal recording. Because the latter is often difficult to distinguish from the ictal phase itself, and might also be informative for seizure detection, we did not attempt to separate true ictal from immediate post-ictal activities during the one minute of ictal recording. The interictal segments were considered as “negative class” segments, while the ictal segments were considered as “positive class” segments.  The immediate pre-ictal minute was not used for training nor for performance evaluation but was used for detecting seizure onsets as explained in section \ref{section:post_proc}. The segmentation of each EEG signal is shown in Figure \ref{fig:signal_cutting}. Each segment is then subdivided in windows of 5 seconds duration with 50\% overlap for data augmentation and centered with respect to the median of the window. This allows to balance training data with each file including 23 negative samples and 23 positive samples. No additional pre-processing steps are performed to ensure a fast online detection application. Although a 50Hz power line noise is present in some EEG signals, we did not want to remove it as it does not affect one class more than another and should not be learned by the CNN model as a decisive factor. The CNN input dimension is then of 1280 time samples (5 seconds x 256Hz) and 4 channels. These data are fed to the network as 2D gray-scale images where the height and width dimensions are the time and channels respectively.

\begin{figure}[h!!]
    \centering
    \includegraphics[width=0.75\textwidth]{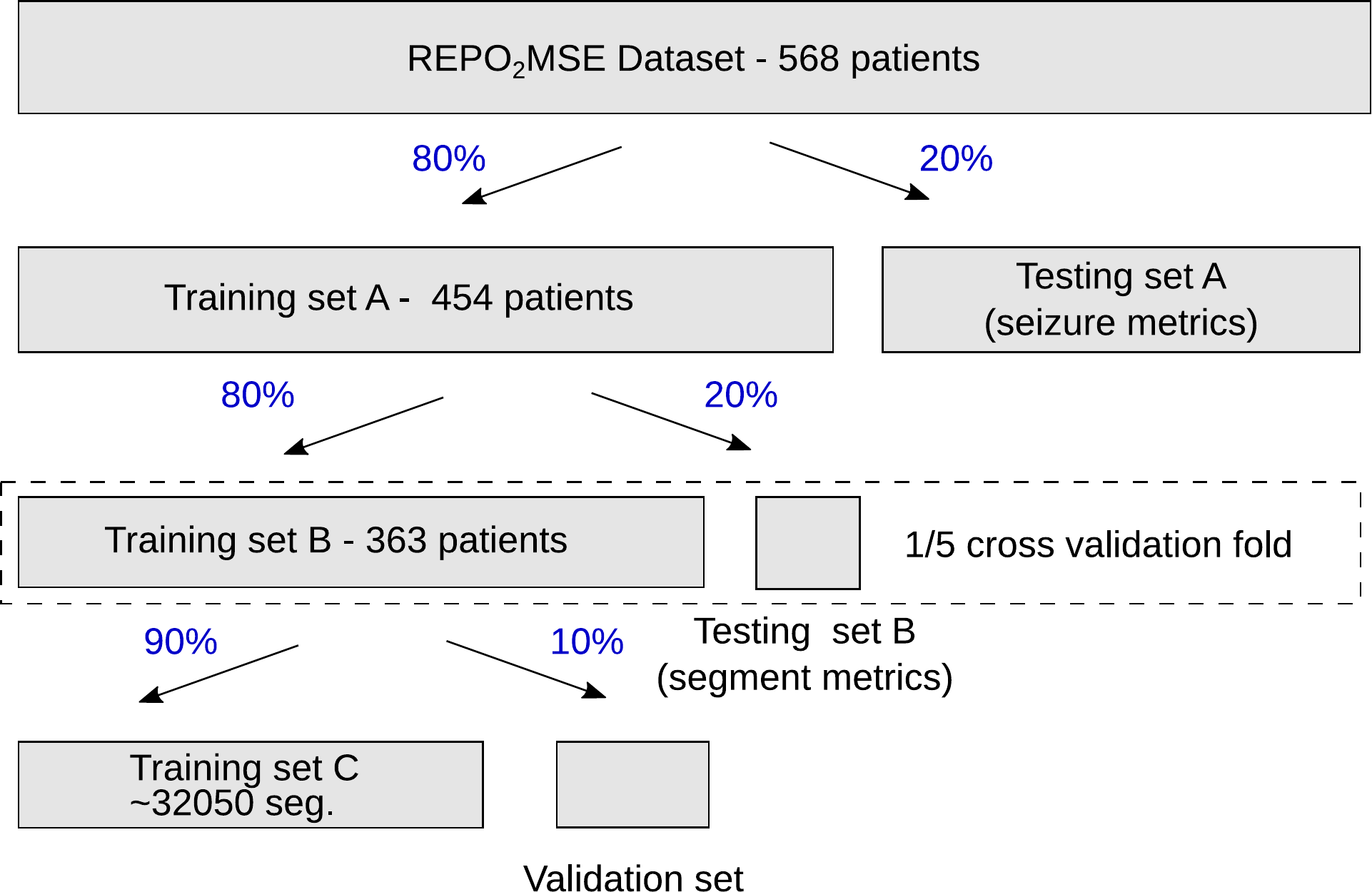}
    \caption{\textbf{Database splitting.} Data from 568 patients are used. 20\% are left out for testing at the seizure-level (test set A). 80\% are used for both cross-validation training (B) to obtain metrics at the segment-level and again for training to obtain metrics at the seizure-level without cross-validation (A). 5-fold CV requires 80\% of training data (B) as the training set. Because we use early stopping, we further select 90\% of the training set (C) as the final training set and leave 10\% as validation metrics during training. All steps except the last ensure data independence at the patient level. The number of 5-seconds segments used for training is not fixed at each fold as not all patients have the same amount of data. 57.6\% of the full database is used for training at each cross-validation fold and 72\% for final training.}
    \label{fig:database}
\end{figure}

The method to split the data between training, validation and testing sets is illustrated in Figure \ref{fig:database}. 80\% of the data are used for training and testing at the segment-level and 20\% for testing at the seizure-level. They are referred to as training and testing sets A, respectively. We perform 5-fold cross-validation (CV) splitting on the patient list with 10\% of validation data at each fold on the training set to obtain metrics at the segment-level and fine-tune hyper-parameters used for detection of the seizure onset. They are referred to as training and testing sets B. Splitting on the patient indices instead of on the short-segments ensures data independence. The split does not correspond exactly to 80\% of the segments as some patients have more seizure files than others resulting in a slight imbalance at each fold. 10\% of training set B is used as validation to monitor the training performance on training set C and implement early-stopping. When training on set A without cross-validation, 10\% of the samples are used for validation.

\subsection{Network architecture}\label{section:architect}

\begin{figure}
    \centering
    \includegraphics[width=0.7\textwidth]{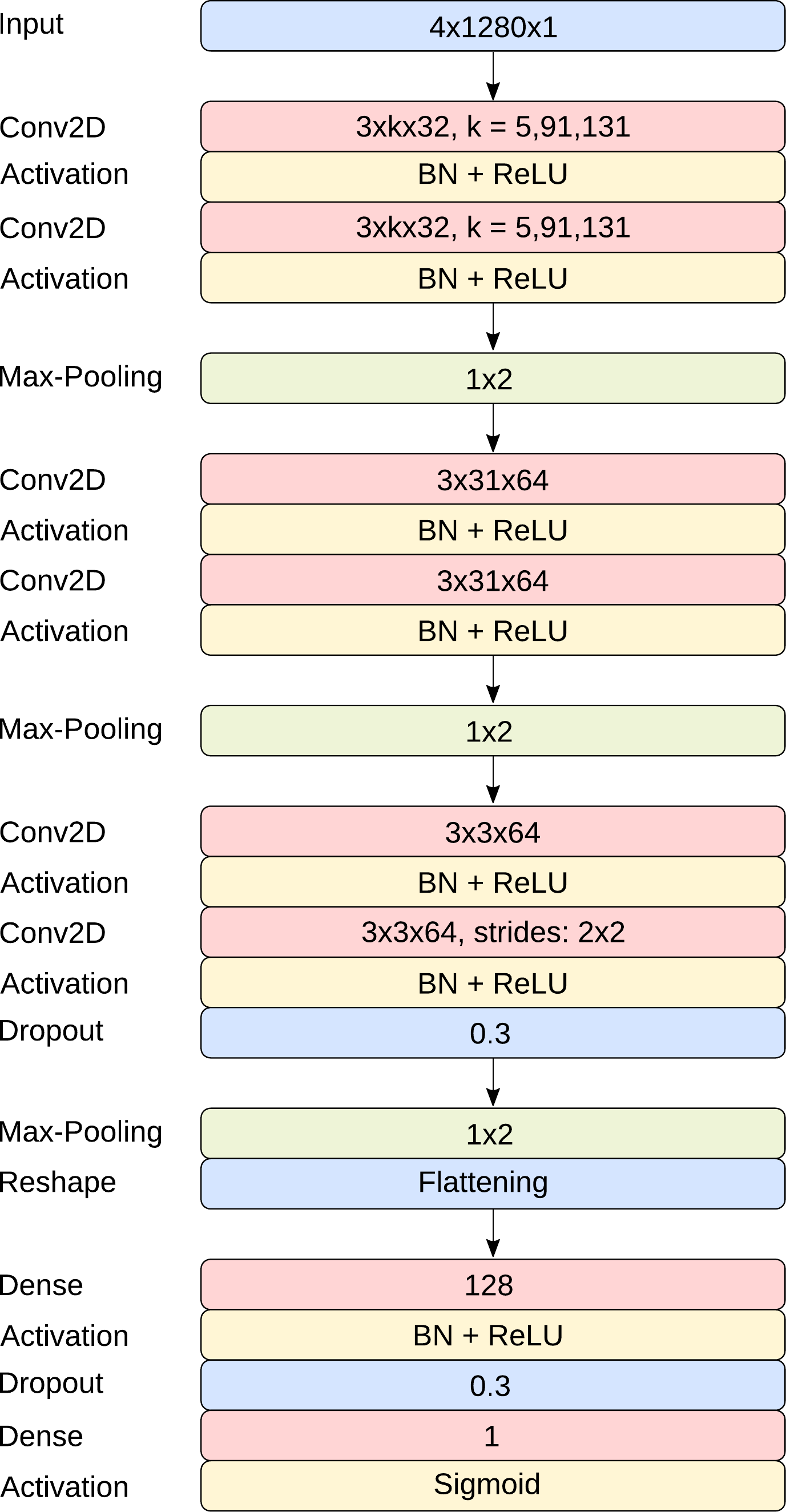}
    \caption{\textbf{Neural network architecture.}}
    \label{fig:network}
\end{figure}

For this study, we decided to develop a CNN architecture, as this type of models have shown promising results in epilepsy detection and classification in different works~\cite{truong_convolutional_2018, ullah_automated_2018, de_aguiar_early_2015}. The model architecture is illustrated in Figure~\ref{fig:network}. It is composed of three blocks of convolutional layers followed by two fully connected (FC) layers with a single output for binary classification (i.e, ictal vs interictal). Each block of convolutional layers consists of two units of a convolutional layer, followed by a Batch Normalization (BN) operation and Rectified Linear Unit (ReLU) activation. Convolutional layers in the first block contains 32 3x$k$ kernels, 64 3x31 kernels in the second block, and 64 3x3 kernels in the last block with a stride of 1x1 in all blocks. To explore the features learned by the model in the first layer, we tested different values of $k$ in the first layer filters. We compare models with kernel size of 3x5, 3x91, and 3x131. In what follows we refer to each one of these models according to the first layer kernel size. As mentioned earlier, these large kernel sizes help the model to extract low frequency content directly from the signal. A kernel of length 131 can capture frequencies as low as 2Hz, while a kernel of length 5 can only explicitly extract high-gamma frequencies in the first two layers. In the latter case, the following 3x31 block can also extract lower frequencies. The temporal sizes of the kernels were chosen arbitrarily and fine-tuned experimentally, but they were kept distinct enough to extract different ranges of frequencies.

A max pooling operation follows each block to reduce dimensionality and to improve temporal invariance of the input. BN is used to re-center the data and to ensure a non-linear ReLU activation, as this has been proven to speed-up training and improve model performance \cite{ioffe_batch_2015}. A 30\% dropout (DO) operation is used before each FC layer and 0.05 weighted kernel and bias regularization are used to avoid overfitting. Models are trained for a maximum of 120 epochs with early-stopping if no improvement of the validation loss is made after 15 epochs. Stochastic Gradient Descend (SGD) optimizer is used with a learning-rate of 0.005. The training was performed on a computing server with 2 AMD EPYC 7551 32-Core Processors and 500~GB of RAM, equipped with an Nvidia Tesla T4 GPU. The models were implemented with Keras using the Tensorflow v1.4 backend. Training on set A takes 99 minutes for 50 epochs, or a mean time of 118 seconds per epoch. 

\subsection{Post-processing} \label{section:post_proc}
We report accuracy, sensitivity, precision, and F1-score both at the segment and seizure-levels. Metrics at the segment-level are reported by concatenating results for each CV-fold while Area Under the receiver operating characteristic Curve (AUC) is averaged across CV folds (testing sets B) \cite{forman_apples--apples_2010}. Results at the seizure-level are reported after a simple training on 80\% of patients (training set A) and testing on the remaining 20\% (testing set A).

Two methods are used in this study to aggregate predictions from different segments in order to detect seizure events. We first define a seizure detection method based on evidence aggregation for the ictal and interictal classes, following a Bayesian approach. By taking the product of consecutive prediction outputs over a sliding window of size $M$, we compute the evidence for the ictal and interictal class respectively. To reduce false positive detection in the interictal part, we consider a window positive if the log-odds of the ictal evidence over the interictal evidence is superior to a threshold $th$. This method assumes statistical independence between each subsequent segment, which is not the case for EEG time-series, but it is still the method of choice since each segment is processed independently by the network. As there is no signal without seizures in the dataset, metrics at the seizure-level are reported according to the negative and positive parts of the signal. A seizure is correctly detected if the onset is detected in the ictal part of the signal and counted as a false positive if detected in the interictal part (see Figure~\ref{fig:signal_cutting}).

We also defined a difference filter method aiming at detecting the transition between the pre-ictal and ictal segments. Previous output probability for a segment at time $t=-M$ is subtracted from the output probability at time $t=0$, where $M$ is the length of the difference filter. To account for values of $M$ up to 23 samples, we also considered the 1 minute of signal before the interictal part so that the false positive rate is not artificially reduced in the interictal part. 75 recordings from 23 patients did not contain enough interictal signal and thus were not considered for testing at the seizure-level. We used the same metric computations in both methods for a fair comparison. The hyper-parameters of each method were optimized on the concatenated cross-validation sets and were used to generate metrics at the seizure-level on the test set A.

%

\subsection{Visualization}\label{section:visualization}
Two visualization methods are employed to understand the decision dynamics of the network on the EEG signal. First, we explore what are the inputs maximizing the first layer kernels. By choosing not to transform the signals into spectrograms, we want to observe if the frequency content of the input EEG is extracted in the first layers and if it relates to the current knowledge in epilepsy. Generation of inputs maximizing the first layer kernels is performed using gradient ascent for 80 epochs and a step size of 0.5 with normalization after each step. Inputs are initialized with random samples in the range [-10$\mu V$;10$\mu V$]. Maximised inputs are sorted according to their descending contribution to a mean loss function after the first ReLU-activation operation. As the resulting signals were sinusoids, we computed their power spectrum with the Welch's method and reported the main frequency components for each channel \cite{welch_use_1967}. 

Even if the latter method allows us to understand frequency components extracted by the first layer kernels, it does not easily convey what are the signal features that led to a specific decision. To visualize the learned features back on the EEG signal, we use the DeepLIFT algorithm. This method compares the output difference between a baseline signal and the current input and propagates this difference back to the input signal referred to as shap values~\cite{shrikumar_features_2017}. We can then read the features of the input EEG that were different from the baseline EEG and led to a decision. The baseline signal corresponds to a zero EEG in all channels, and since it might appear similar to the negative class signal to the network, we only display differences between the positive class and the baseline value. 

\section{Experimental results}

In this section, we first discuss the detection performance of the three models both at the segment and seizure-levels. We then explore the results of the maximized inputs and of the DeepLIFT visualization method.

\subsection{Model performance}

\begin{table}[b!]
\centering
\begin{tabular}{|l|ll|ll|ll|}
\hline
\textbf{Model} &  \multicolumn{2}{c|}{\textbf{3x91}} & \multicolumn{2}{c|}{\textbf{3x131}} &
  \multicolumn{2}{c|}{\textbf{3x5}} \\ \hline
Threshold & 0.150 & 0.850 & 0.150 & 0.850 & 0.150 & 0.850 \\ \hline
Accuracy  & 0.705 & 0.686 & \textbf{0.711} & 0.644 & 0.657 & 0.690 \\ \hline
Sensitivity & 0.930 & 0.405 & 0.914 & 0.307 & \textbf{0.962} & 0.416 \\ \hline
Precision & 0.481 & 0.966 & 0.508 & \textbf{0.981} & 0.353 & 0.964 \\ \hline
F1-score & 0.759 & 0.563 & \textbf{0.760} & 0.463 & 0.737 & 0.573 \\ \hline
AUC & \multicolumn{2}{l|}{0.866 $\pm$ 0.02} &
  \multicolumn{2}{l|}{0.867 $\pm$ 0.022} &
  \multicolumn{2}{l|}{0.866 $\pm$ 0.026} \\ \hline
\end{tabular}
\caption{\textbf{Metrics at the segment-level.} Accuracy, sensitivity, precision, and F1-score are reported by concatenating CV outputs (test sets B). Network output is converted to binary classification according to two decision thresholds chosen arbitrarily to explore performance at low- and high-sensitivity. Overall, 3x5 model exhibits the highest sensitivity and 3x131 the best precision. Averaging AUC across folds shows that all three models perform equally.}
\label{table:metrics_segment}
\end{table}

\begin{figure}[h]
    \centering
    \includegraphics[width=\textwidth]{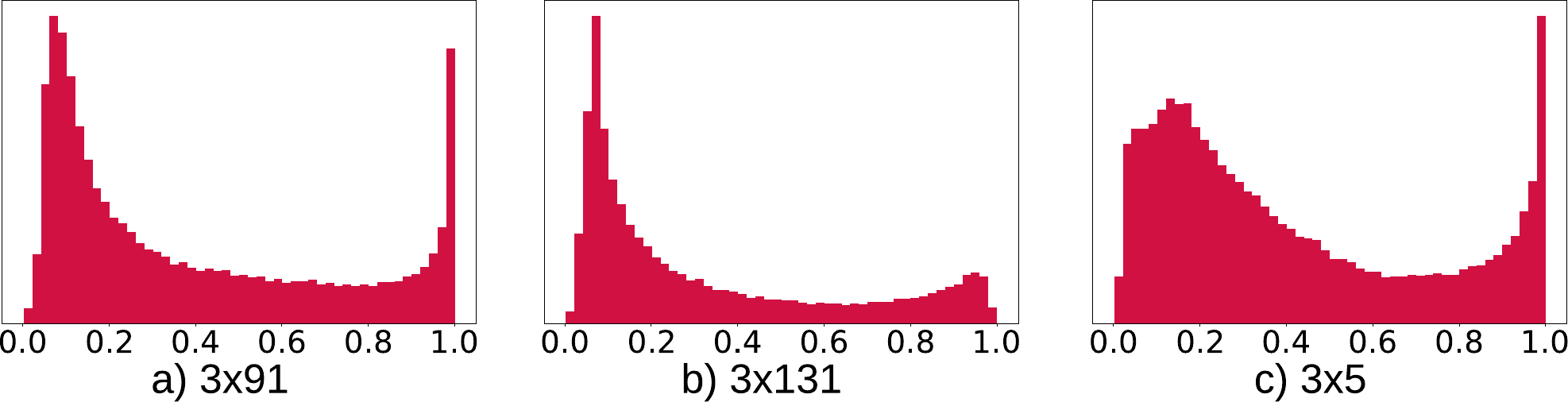}
    \caption{\textbf{Distribution of output probabilities}. Output probabilities of each CNN are plotted as a histogram. 3x5 model has probabilities shifted to the right, suggesting a higher sensitivity. On the contrary 3x131 model has few outputs close to one indicating a potentially lower sensitivity.} 
    \label{fig:output_distribution}
\end{figure}

Table \ref{table:metrics_segment} shows the classification performance metrics at the segment-level for the 3x91, 3x131 and 3x5 models. We applied two different decision thresholds: 0.15 and 0.85 for high and low sensitivity respectively. Figure \ref{fig:output_distribution} displays the prediction polarisation at the output neuron. Comparing Table \ref{table:metrics_segment} and Figure \ref{fig:output_distribution} gives thorough information on networks dynamics at the segment-level.
The 3x131 model yields the best F1-score of 0.76 at the low threshold and is the model with output probabilities that are the most shifted towards the negative class. On the contrary, the 3x5 model has output probabilities shifted towards the ictal class and the best F1-score of 0.577 at a high decision threshold. In both cases, the output probability distribution matches network performance metrics. The 3x5 model can keep relatively high sensitivity at a high threshold because the distribution is more shifted towards the positive class than for the other models and the opposite is observed for the 3x131 model. The 3x91 model is the most balanced with F1-scores of 0.759 and 0.566 for decision thresholds of 0.15 and 0.85 respectively. All models have an AUC of 0.87. Performance at the segment-level is in line with the aforementioned studies using DL. Direct comparison is however not possible as this study constitutes the first DL work on the target dataset.

This summarizes the performance of the models at the segment-level. However, the performance of the seizure detection method depends on complete EEG signals and not only on classification of short segments. Since the detection of seizure patterns does not require that all sub-segments are classified correctly, a careful design of the post-processing method can potentially increase the performance at the seizure-level.

\subsubsection{Evidence aggregation for seizure classification}

\begin{figure}
    \centering
    \includegraphics[width=\textwidth]{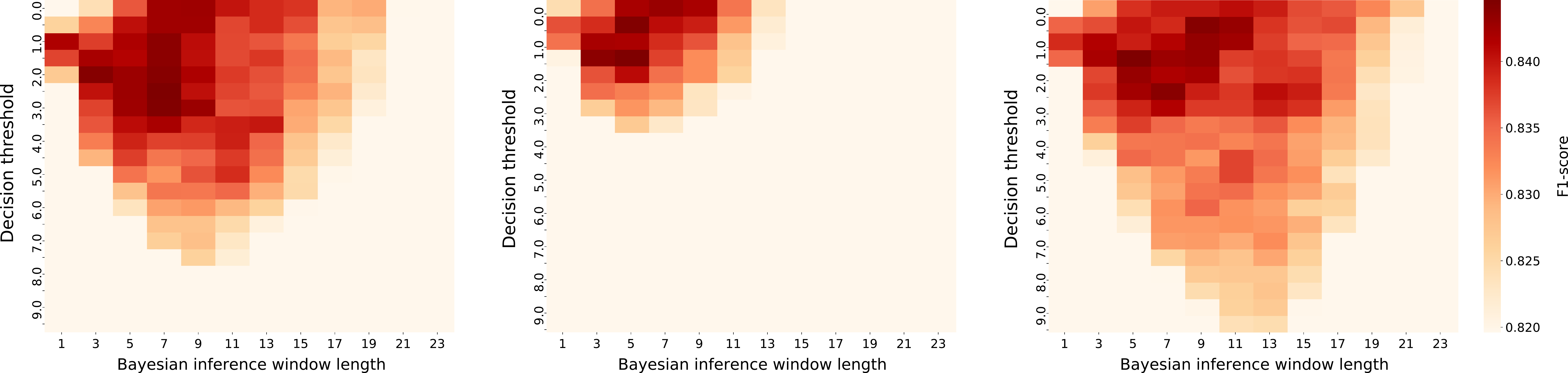}
    \caption{\textbf{Hyper-parameters space - Bayesian approach.} a. 3x91, b. 3x131, c. 3x5. F1 scores are smoothly distributed in the hyper-parameter space of every model, showing a low risk of overfitting. The 3x5 model has optimal hyper-parameters shifted towards high log-odds as a consequence of a high sensitivity at the segment-level. The opposite is observed for the 3x131 model.}
    \label{fig:heatmap_cv_bay}
\end{figure}

A Bayesian approach allows to process the $W$ previously classified segments together to give an output decision at a given time at the seizure-level. To apply this method, we optimized two hyper-parameters on the concatenated CV folds test sets B. The optimal window size $W^*$ and optimal threshold $th^*$ are greedily selected by maximizing the F1-score. Grid selection is displayed in Figure \ref{fig:heatmap_cv_bay} along with selected hyper-parameter values in Table \ref{table:optimization_bayesian}. These parameters are then used to compute metrics at the seizure-level on the left out test set (test set A). Only the ictal and interictal parts of each signal are used to assess the performance of the models.

Accuracy, sensitivity, precision, and F1-score are reported in Table \ref{table:metrics_signal_bayesian}. The 3x131 model has the highest sensitivity with 89.5\% of the test seizures detected, while the 3x91 has the lowest one with a value of 83.4\%. The 3x5 model has the best F1-score of 0.853 and the highest precision, with less than 9.82\% of false positives. Middle column of Figure \ref{fig:post_process} shows the output of the evidence aggregation method applied to all test signals. We can observe that the behavior of each model at the seizure-level is similar, showing that optimizing hyper-parameters compensated the differences at the segment-level. For each model, falsely classified segments mostly correspond to signals being classified either as fully ictal or fully interictal with no apparent transition. These signals are consistent across each model showing that part of the seizure signals could never be properly detected with any of the proposed models.

If we consider the specific optimum values for the hyperparameters, we can see that the optimal analysis window is around 15-20 seconds, which also corresponds to the typical length of screen windows used by neurologists when reading EEG, and suggest that the time scale of the relevant phenomena is very similar for the DL models and for the human experts. Regarding the evidence threshold, a log-odds value between 1.5 and 2.5 corresponds to an aggregated probability approximately between 0.8 and 0.9, meaning that high evidence is required to predict a positive output.

\begin{table}
\centering
\begin{tabular}{|l|l|l|}
\hline
\textbf{Parameters} & $W^*$ & $th^*$  \\ \hline
3x5   & 5 & 1.5 \\ \hline
3x91  & 7 & 2.5 \\ \hline
3x131 & 5 & 1.5 \\ \hline
\end{tabular}
\caption{\textbf{Optimized hyper-parameters for the Bayesian approach for classification at the seizure-level.}}
\label{table:optimization_bayesian}
\end{table}

\begin{table}
\centering
\begin{tabular}{|l|l|l|l|}
\hline
\textbf{Model} &  \multicolumn{1}{c|}{\textbf{3x91}} & \multicolumn{1}{c|}{\textbf{3x131}} & \multicolumn{1}{c|}{\textbf{3x5}} \\ \hline
Sensitivity & 0.834          & \textbf{0.895} & 0.886          \\ \hline
Precision   & 0.825          & 0.795          & \textbf{0.808}\\ \hline
Accuracy    & 0.83          & 0.845           & \textbf{0.847} \\ \hline
F1-score    & 0.83          & 0.852           & \textbf{0.853}  \\ \hline
\end{tabular}
\caption{\textbf{Metrics at the seizure-level - Bayesian approach.} }
\label{table:metrics_signal_bayesian}
\end{table}

\subsubsection{Difference filter}

The second implemented method for seizure classification from the outputs at the segment-level aims at detecting the transition between the pre-ictal and the interictal portions of each signal. Every 2.5 seconds, the output decision is based on predictions at windows $T$ and $T-M$.

\begin{figure}
    \centering
    \includegraphics[width=\textwidth]{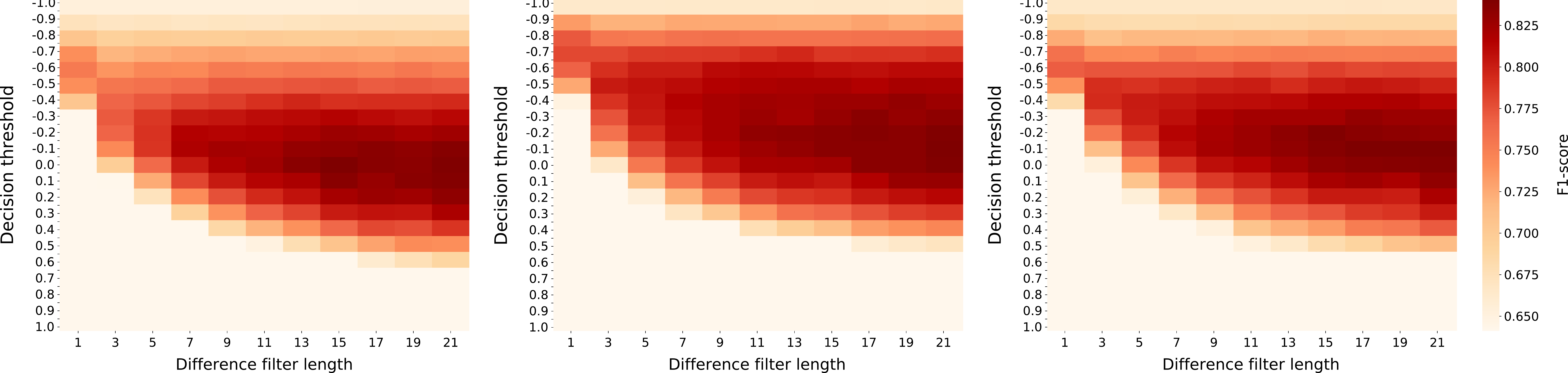}
    \caption{\textbf{Hyper-parameters space - Difference Filter.} a. 3x91, b. 3x131, c. 3x5. Hyper-parameters space is smooth reducing the risk of overfitting. 3x131 model space is shifted towards lower threshold values. All models seem to have an optimal size of filter beyond 23. This factor is limited by the short time of recording before seizure onset label.}
    \label{fig:heatmap_cv_diff}
\end{figure}

This method also requires the optimization of two hyper-parameters. $M$ is the distance between two samples and $th_d$ the threshold above which a difference between two samples is considered as a seizure onset indicator. As for the Bayesian approach, hyper-parameters are greedily selected on the concatenated CV sets and used to compute metrics at the seizure-level on the test set A. Optimal $M^*$ and $th^*$ are displayed in Table \ref{table:optimization_difference} and the F1-score grids in Figure \ref{fig:heatmap_cv_diff}. Hyper-parameter spaces are similar in shapes for every model, with a shift towards lower thresholds for the 3x131 model.

Table~\ref{table:metrics_signal_difference} shows the metrics at the seizure-level for the difference filter method. The extended interictal and pre-ictal portions of the signals are not taken into account for metrics computation. They are required for keeping a true false-positive rate in the interictal part and continuous processing, respectively. The 3x131 model performs best with an F1-score of 0.873 and 90\% of the seizures detected, while the 3x91 model generates the lowest false positive rate in the interictal part with a precision of 0.891. This can be explained by a higher $th^*$ and shorter $M^*$. 

\begin{table}[]
\centering
\begin{tabular}{|l|l|l|}
\hline
\textbf{Parameters} & $M^*$ & $th^*$  \\ \hline
3x5   & 17 & 0.45 \\ \hline
3x91  & 15 & 0.5 \\ \hline
3x131 & 21 & 0.45 \\ \hline
\end{tabular}
\caption{\textbf{Hyper-parameters optimization - Difference filter}}
\label{table:optimization_difference}
\end{table}

\begin{table}[]
\centering
\begin{tabular}{|l|l|l|l|}
\hline
\textbf{Model} &  \multicolumn{1}{c|}{\textbf{3x91}} & \multicolumn{1}{c|}{\textbf{3x131}} & \multicolumn{1}{c|}{\textbf{3x5}} \\ \hline
Sensitivity & 0.817             & 0.904            & \textbf{0.908}          \\ \hline
Precision   & \textbf{0.891}    & 0.834            & 0.76           \\ \hline
Accuracy    & 0.854             & \textbf{0.869}   & 0.834          \\ \hline
F1-score    & 0.848             & \textbf{0.873}   & 0.846          \\ \hline
\end{tabular}
\caption{\textbf{Metrics at the seizure-level - Difference Filter.}}
\label{table:metrics_signal_difference}
\end{table}

\subsection{Methods comparison}
For both methods, performance at the seizure-level is higher than at the segment-level. At the segment-level, the F1-scores never exceeded 0.76, while it reaches 0.873 for the 3x131 model with the difference filter method. This was expected as some short-segments were not carrying explicit ictal features for short seizures and detection at the seizure-level requires only one window to be positive in the ictal class. Performance differences at the segment-level are not conserved at the seizure-level. Indeed, the 3x131 model was yielding the highest precision and the 3x5 model the highest sensitivity, while it is the opposite at the seizure-level for the Bayesian approach. This effect is less pronounced for the difference filter. To verify if this is a result of hyper-parameter optimization, we computed the metrics for both methods with fixed sets of hyper-parameters. Results are shown in Tables \ref{table:metrics_signal_difference_comparison_baye} and \ref{table:metrics_signal_difference_comparison_diff}. They reveal the previous dynamic observed at the segment-level where 3x5 model was yielding the highest sensitivity and 3x131 model the highest precision. Therefore, hyper-parameter optimization tends to make methods converge to a common behavior by compensating differences at the seizure-level.

\begin{figure}
    \centering
    \includegraphics[width=\textwidth]{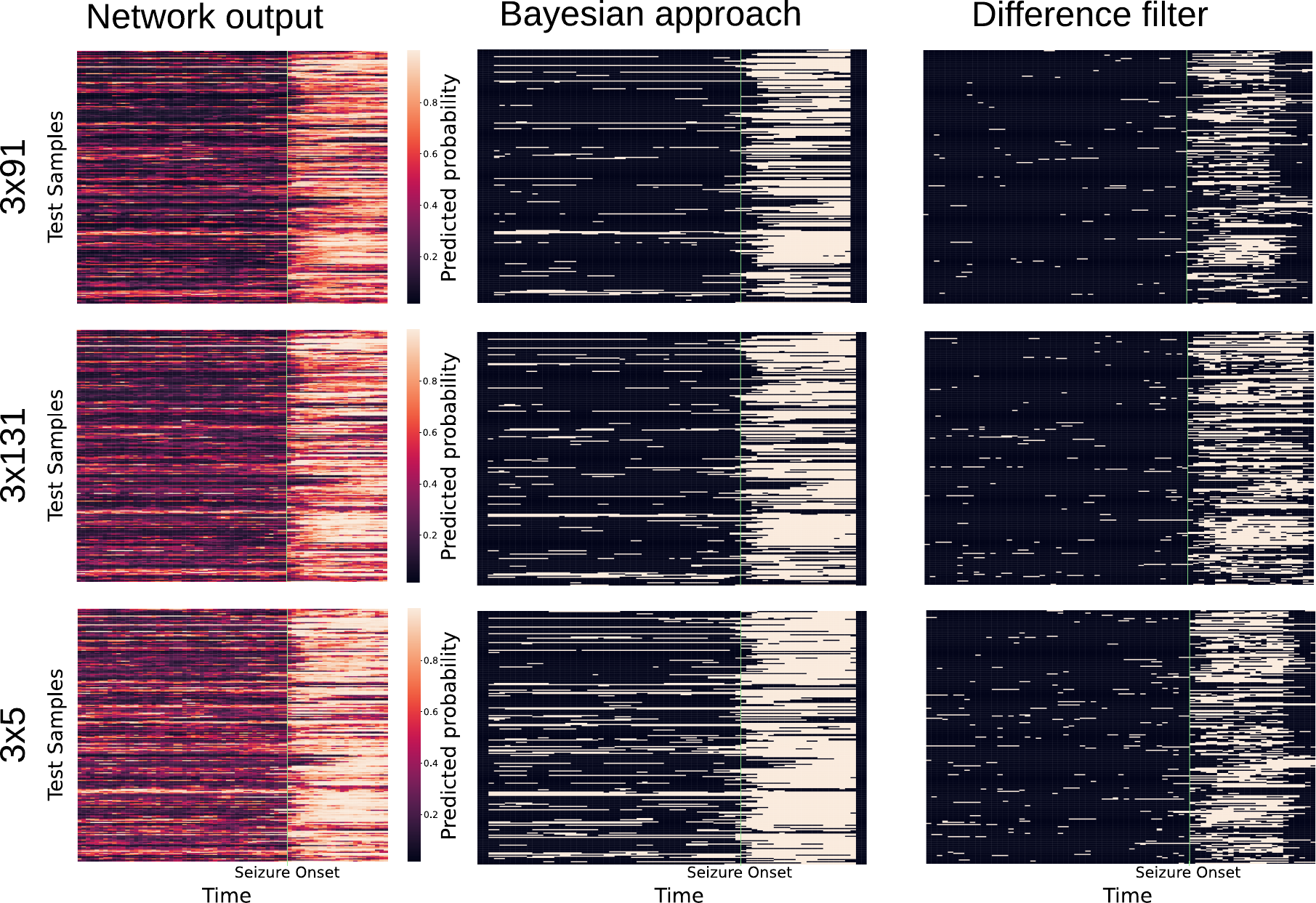}
    \caption{\textbf{Comparison of post-processing methods.} Network outputs (left column) are presented as heatmaps with values ranging from 0 to 1 for the negative and positive class respectively. The raw outputs are then transformed by the Bayesian approach (middle column) or the difference filter method (right column). The Bayesian approach closely resembles the network output, with high sensitivity but also a higher false positive rate. On the contrary, the difference filter is more focused on the seizure onset and has a lower false positive rate.}
    \label{fig:post_process}
\end{figure}

Figure \ref{fig:post_process} compares the outputs of both methods. The left column represents the continuous probability outputs of the model while the middle and right columns represent the post-processed binary outputs for the Bayesian approach and difference filter respectively. Each line of the heatmap corresponds to one of the 229 signals in the test set A. The Bayesian approach often leads to higher sensitivity and longer stretches of ictal detection. False positives in the interictal part are reduced for the difference method, as this method can detect drifts in output probabilities stronger than $th^*$ for signals with probabilities constantly above 0.5. This explains why some signals are considered as fully ictal in the Bayesian approach while not being detected as such by the difference method.

\begin{table}
\centering
\begin{tabular}{|l|ll|ll|ll|}
\hline
\textbf{Model} &  \multicolumn{2}{c|}{\textbf{3x91}} & \multicolumn{2}{c|}{\textbf{3x131}} & \multicolumn{2}{c|}{\textbf{3x5}} \\ \hline
Threshold   & 0.5         & 3.0         & 0.5          & 3.0         & 0.5         & 3.0        \\ \hline
Sensitivity & 0.904       & 0.817       & 0.895        & 0.865       & 0.974       & 0.93       \\ \hline
Precision   & 0.803       & 0.847       & 0.782        & 0.852       & 0.664       & 0.729      \\ \hline
Accuracy    & 0.854       & 0.832       & 0.838        & 0.858       & 0.819       & 0.83       \\ \hline
F1 score    & 0.861       & 0.829       & 0.847        & 0.859       & 0.843       & 0.845      \\ \hline
\end{tabular}
\caption{\textbf{Metrics at the seizure-level with $W=7$ - Bayesian approach} 3x91 model performs the best for low threshold and 3x131 is the best classifier at a high threshold. This last result is rather surprising as the observations are opposed to the difference filter method. Antithetic dynamics of 3x131 and 3x5 models is still preserved.}
\label{table:metrics_signal_difference_comparison_diff}
\end{table}

\begin{table}
\centering
\begin{tabular}{|l|ll|ll|ll|}
\hline
\textbf{Model} &  \multicolumn{2}{c|}{\textbf{3x91}} & \multicolumn{2}{c|}{\textbf{3x131}} & \multicolumn{2}{c|}{\textbf{3x5}} \\ \hline
Threshold   & 0.4         & 0.6         & 0.4          & 0.6         & 0.4         & 0.6        \\ \hline
Sensitivity & 0.904       & 0.729       & 0.908        & 0.764       & 0.93        & 0.795      \\ \hline
Precision   & 0.764       & 0.943       & 0.782        & 0.948       & 0.721       & 0.913      \\ \hline
Accuracy    & 0.834       & 0.836       & 0.845        & 0.856       & 0.825       & 0.854      \\ \hline
F1 score    & 0.845       & 0.817       & 0.854        & 0.841       & 0.842       & 0.845      \\ \hline
\end{tabular}
\caption{\textbf{Metrics at the seizure-level with $M=19$ - Difference Filter.} 3x131 model performs the best at a low threshold and 3x5 at a high threshold. This highlights again the opposite performance of both models.}
\label{table:metrics_signal_difference_comparison_baye}
\end{table}

\subsection{Network interpretability}

\subsubsection{Maximized inputs}
\begin{figure}
    \centering
    \includegraphics[width=\textwidth]{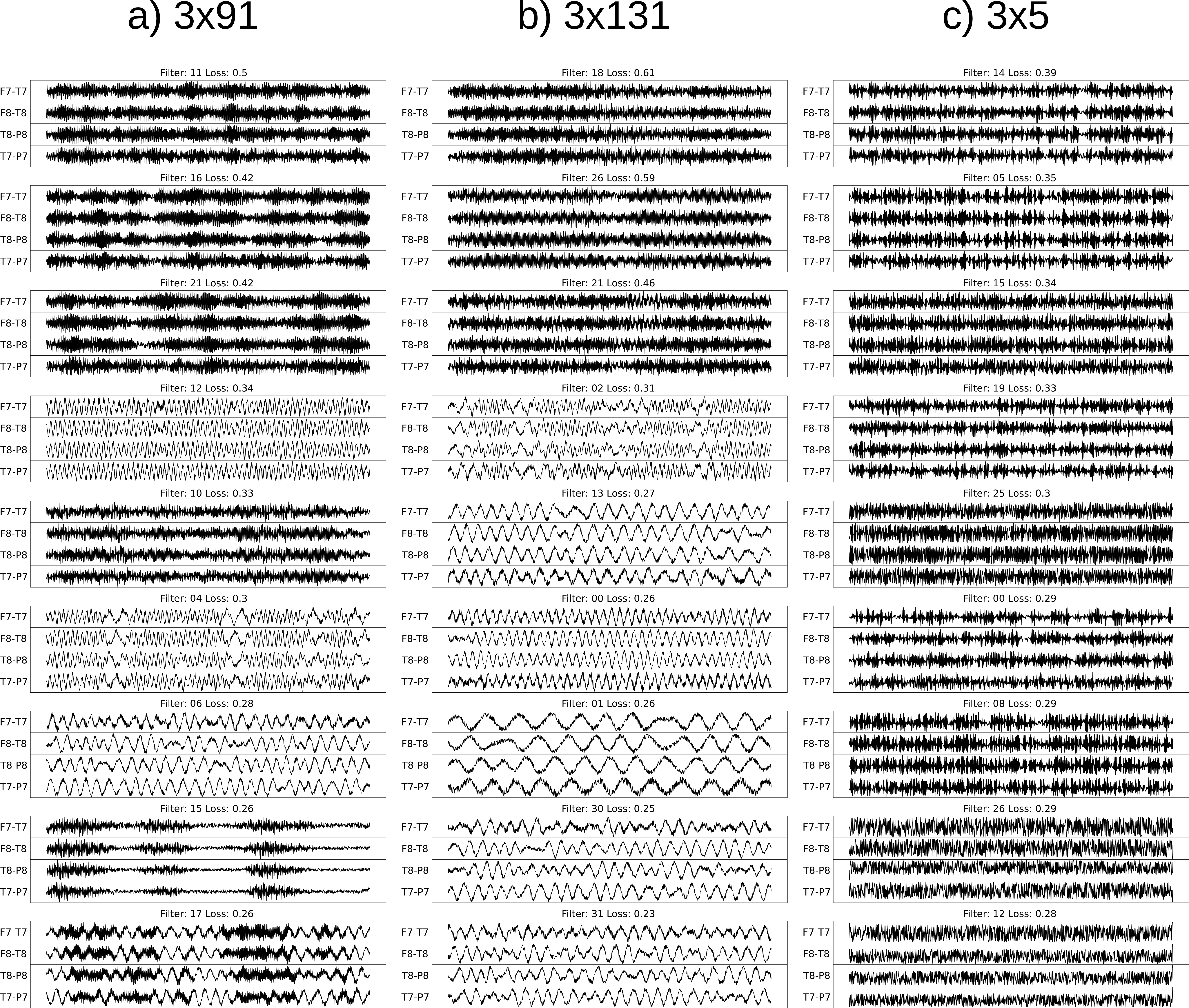}
    \caption{\textbf{Most significant maximized inputs for first layer kernels with low initial amplitude.} Maximized inputs are sorted according to their contribution to a custom loss function after the last gradient ascent step. 3x5 model shows only high frequency detecting kernels on the first layer. Indeed, the kernel sizes can only detect 50Hz frequencies and above with a sampling frequency of 256Hz. 3x131 and 3x91 model both have lower frequency detecting filters. High frequencies detecting filters elicit the strongest response after the first layer in all models.}
    \label{fig:max_input_filters}
\end{figure}

Gradient ascent was performed on the first layer kernels to elucidate their role for feature extraction. The 9 maximized inputs out of 32 filters eliciting the strongest activation for each model are shown in Figure \ref{fig:max_input_filters}. We can observe distinct sinusoidal patterns for 3x91 and 3x131 models after the third maximized inputs. This suggests that first layer kernels play a role in extracting characteristic frequency components in the input EEG and confirms our hypothesis for choosing large kernel sizes. Synchronicity with a small phase shift between channels of maximized inputs can also be observed in most examples. This suggests that spatial correlation can also be an important feature learned by the model. High-frequency components are present in maximized inputs eliciting the strongest activation response suggesting that low frequencies are not the most important features to be extracted in the first layers.

Despite this method being able to highlight important frequency components of the first layer, it does not inform if a given component contributes to the ictal or interictal class. A strong activation at the first layer only suggests that the frequency is important to classify a segment as decision happens at the last sigmoid activation. Accordingly, maximized inputs are fed back to the network and probability at the output is kept. From this probability, we may then infer that a frequency component (or a combination of them) is associated to the ictal class if it has an output probability close to 1, and to the negative class otherwise. Tables \ref{table:3x91_main_freq} and \ref{table:3x131_main_freq} display the main frequency components along with the associated network output probability. We did not report the main frequencies for the 3x5 model as the power spectrum was broad and no singular frequencies could be identified. For 3x131 and 3x91 models, the majority of maximized inputs contain frequencies in the alpha, beta, and low-gamma bands. However, maximized inputs eliciting the strongest activation response contain high-gamma frequencies between 70 and 100 Hz. Those filters also lead to strong activation of the ictal class. Results highlight that most maximized inputs contribute to the ictal class. However, some frequency components are associated with activation of the interictal class. Indeed, maximized inputs containing frequency components around 8Hz seem to be an indicator of the interictal class. This is in agreement with the alpha band being associated with the resting activity of brain waves and therefore a characteristic of interictal activity. 

\begin{figure}
    \centering
    \includegraphics[width=\textwidth]{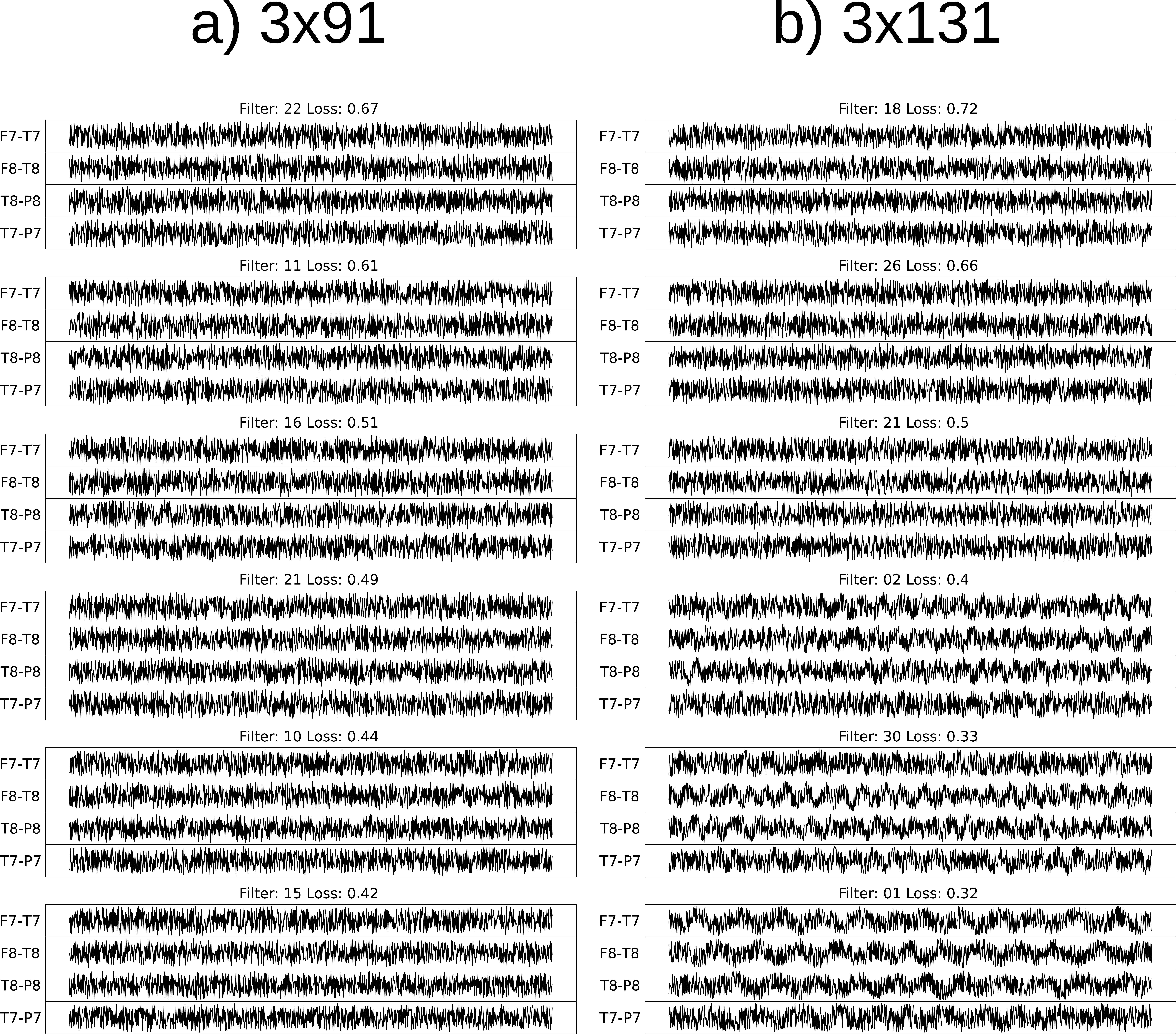}
    \caption{\textbf{Most significant maximized inputs for first layer kernels with high initial amplitude.} Maximized inputs are now initialized with random samples with a magnitude up to 100$\mu V$. High amplitude initialization leads to amplification of high frequencies and stronger activation.}
    \label{fig:max_input_filters_high_amp}
\end{figure}

To assess if high-gamma frequency bands in the maximized inputs are features of the ictal phase or an effect of the gradient ascent method, we raised the input range amplitude to 100~$\mu V$ instead of 10~$\mu V$. Figure \ref{fig:max_input_filters_high_amp} and Tables \ref{table:3x91_main_freq_amp} and \ref{table:3x131_main_freq_amp} now show a much higher proportion of high frequency components in the maximized input in the 3x131 and 3x91 models. This result suggests high frequency components are important for extracting the high amplitude information in the signal. Indeed, the convolution of a high frequency filter with high amplitude peaks on the EEG signal will yield output with high energy potentially conserved after ReLU activation. Moreover, one can observe that output probabilities are all close to 1 in this case. Maximized inputs containing 8Hz frequency components still elicit the lowest output probability although now being in favor of the ictal class. Therefore we can conclude that a large amplitude is a distinctive feature of the ictal class in all three models, and may override other features that were previously indicators of the interictal class.

\begin{table}
\centering
\begin{tabular}{lllllrr}
\toprule
\textbf{Filter idx} &        \textbf{F7-T7} &        \textbf{F8-T8} &        \textbf{T8-P8} &     \textbf{T7-P7} &      \textbf{pred} &      \textbf{loss} \\
\midrule
11 &      [97] &      [97] &      [97] &      [97] &  0.999211 &  0.496292 \\
16 &      [97] &      [97] &      [97] &      [98] &  0.998845 &  0.424636 \\
21 &      [97] &      [97] &      [97] &      [97] &  0.998827 &  0.417835 \\
12 &      [14] &      [14] &      [14] &      [14] &  0.989898 &  0.336636 \\
10 &  [72, 97] &  [72, 97] &  [72, 97] &  [72, 97] &  0.996634 &  0.331971 \\
4  &   [14, 4] &   [14, 4] &   [14, 4] &   [14, 4] &  0.994263 &  0.299618 \\
6  &       [5] &       [5] &       [6] &       [6] &  0.993710 &  0.279007 \\
15 &  [72, 97] &  [72, 97] &  [72, 97] &  [72, 97] &  0.673070 &  0.264437 \\
17 &   [97, 5] &   [97, 5] &   [97, 5] &   [5, 97] &  0.999591 &  0.263853 \\
{[...]} &        {[...]} &      {[...]}   &        {[...]} &           {[...]} &        {[...]} &        {[...]} \\
3  &       [8] &       [8] &       [8] &       [8] &  0.092222 &  0.245886 \\
\bottomrule
\end{tabular}
\caption{\textbf{Main frequency components of maximized inputs with low initial amplitude - 3x91.} The top three maximized inputs contain high frequency components leading to strong activation of the ictal class. Only the 8Hz component is associated with activation of the interictal class.}
\label{table:3x91_main_freq}
\end{table}

\begin{table}
\centering
\begin{tabular}{lllllrr}
\toprule
\textbf{Filter idx} &        \textbf{F7-T7} &        \textbf{F8-T8} &        \textbf{T8-P8} &     \textbf{T7-P7} &      \textbf{pred} &      \textbf{loss} \\
\midrule
18 &  [98, 72] &     [98, 72] &     [98, 72] &  [98, 72] &  0.998613 &  0.605722 \\
26 &  [72, 98] &     [72, 98] &     [72, 98] &      [72] &  0.981446 &  0.593120 \\
21 &      [98] &     [98, 15] &         [98] &      [99] &  0.999239 &  0.460199 \\
2  &   [14, 5] &      [15, 5] &      [15, 5] &   [15, 5] &  0.998223 &  0.313937 \\
13 &       [5] &          [5] &          [5] &       [5] &  0.668976 &  0.265224 \\
0  &       [8] &          [8] &          [8] &       [8] &  0.050489 &  0.258222 \\
1  &       [2] &          [2] &          [2] &       [2] &  0.580311 &  0.258116 \\
30 &       [5] &          [5] &          [5] &       [5] &  0.993062 &  0.246212 \\
31 &       [5] &          [5] &          [5] &       [5] &  0.997932 &  0.233605 \\
\bottomrule
\end{tabular}
\caption{\textbf{Main frequency components of maximized inputs with low initial amplitude - 3x131.} The first three maximized inputs contain high frequency components. As for 3x91 model, 8Hz components are associated with activation of the interictal class.}
\label{table:3x131_main_freq}
\end{table}

\begin{table}
\centering
\begin{tabular}{lllllrr}
\toprule
\textbf{Filter idx} &        \textbf{F7-T7} &        \textbf{F8-T8} &        \textbf{T8-P8} &     \textbf{T7-P7} &      \textbf{pred} &      \textbf{loss} \\
\midrule
22 &      [96] &      [96] &      [97] &      [97] &  0.999452 &  0.668545 \\
11 &      [97] &      [97] &      [97] &      [97] &  0.999715 &  0.606585 \\
16 &      [97] &      [96] &      [97] &      [97] &  0.999315 &  0.514086 \\
21 &      [97] &      [97] &      [97] &      [97] &  0.999397 &  0.493807 \\
10 &  [72, 97] &  [97, 73] &  [97, 72] &  [72, 97] &  0.999416 &  0.444842 \\
15 &      [97] &  [97, 72] &  [96, 72] &  [72, 98] &  0.999123 &  0.423038 \\
12 &      [14] &      [14] &      [14] &      [14] &  0.998754 &  0.400928 \\
4  &      [15] &   [15, 4] &   [14, 4] &        [] &  0.999290 &  0.375046 \\
6  &       [5] &       [4] &       [5] &       [6] &  0.999461 &  0.355838 \\
{[...]} &        {[...]} &      {[...]}   &        {[...]} &           {[...]} &        {[...]} &        {[...]} \\
3  &       [8] &       [8] &       [8] &       [8] &  0.904117 &  0.324952 \\
\bottomrule
\end{tabular}
\caption{\textbf{Main frequency components of maximized inputs with high initial amplitude - 3x91.} Most maximized inputs carry high-gamma frequency components. All associated probabilities are close to 1 and components around 8Hz lead to the lowest prediction probability.}
\label{table:3x91_main_freq_amp}
\end{table}

\begin{table}
\centering
\begin{tabular}{lllllrr}
\toprule
\textbf{Filter idx} &        \textbf{F7-T7} &        \textbf{F8-T8} &        \textbf{T8-P8} &     \textbf{T7-P7} &      \textbf{pred} &      \textbf{loss} \\
\midrule
18 &      [98] &     [98, 72] &         [98, 71] &      [98] &  0.999737 &  0.716940 \\
26 &      [72] &     [72, 98] &         [71, 98] &      [71] &  0.999527 &  0.661337 \\
21 &  [99, 16] &     [99, 15] &         [99, 15] &  [99, 16] &  0.999792 &  0.504515 \\
2  &       [5] &      [4, 15] &          [4, 15] &        [] &  0.999709 &  0.395584 \\
30 &       [5] &          [5] &              [5] &       [5] &  0.999484 &  0.331554 \\
1  &       [2] &          [3] &              [3] &       [2] &  0.993785 &  0.317364 \\
13 &       [4] &          [4] &              [5] &       [5] &  0.999157 &  0.316492 \\
31 &       [6] &          [6] &           [5, 7] &       [5] &  0.999735 &  0.313957 \\
0  &       [8] &          [8] &              [8] &       [8] &  0.946293 &  0.297796 \\
\bottomrule
\end{tabular}
\caption{\textbf{Main frequency components of maximized inputs with high initial amplitude - 3x131.} All the maximized inputs looking noisier, this model does not have a stronger proportion of high-gamma frequency components. As for the 3x91 model, most maximized inputs lead to strong activation of the ictal class. 8Hz components are associated with the lowest output probability.}
\label{table:3x131_main_freq_amp}
\end{table}

\clearpage
\subsection{Inference Visualization}

\begin{figure}[H]
    \centering
    \includegraphics[width = \textwidth]{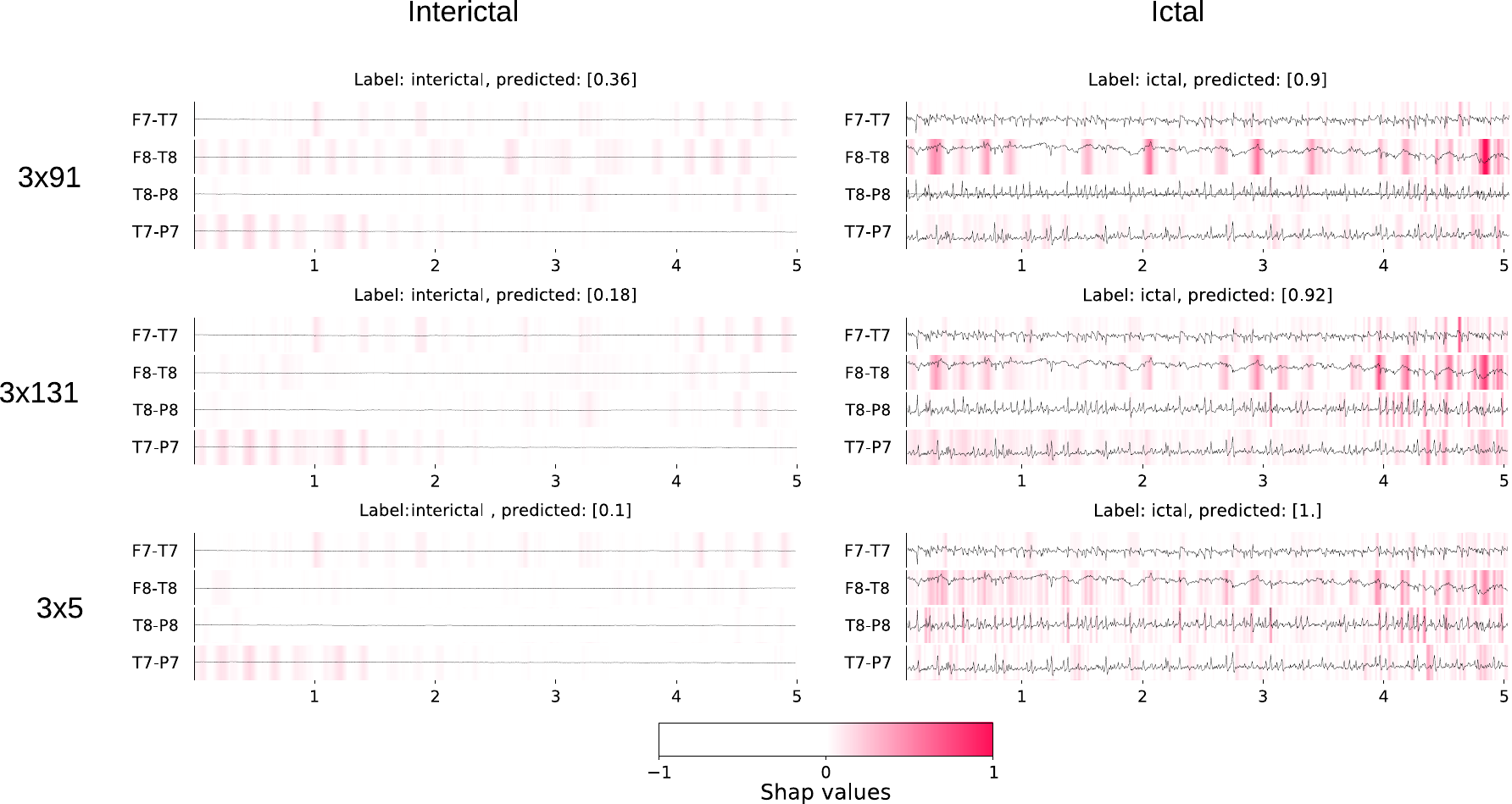}
    \caption{\textbf{Comparison of approach visualization across models and time}. Both windows are taken from the same signal and fed to all three networks. Shap values are then computed and over-layed on the EEG signal after smoothing. Red bands correspond to cluster of positive shap values and indicate decisive ictal features.}
    \label{fig:shap_compar}
\end{figure}

We highlight input features characteristic of the ictal class by overlaying the matrix of shap values on the EEG signal. The normalized matrix contains the back-propagated prediction differences at the output neuron with a  baseline signal. A positive shap value indicates an input data point that led to a positive difference output and therefore a contribution to the positive class. Examples of an ictal and interictal segments are reported in Figure \ref{fig:shap_compar} for all three models. As expected the ictal portion contains a higher proportion of positive shap values than the interictal part. Due to the visual nature of this method, we only report qualitative results, preventing precise comparison between models. On the example presented, learned features are consistent across models. The 3x5 model has a greater number of sharper bands of positive shap values, which could be expected given a higher sensitivity and lower kernel sizes in the time dimension. It also shows that amplitude is not sufficient to classify a segment as ictal since drifts in the signal are not detected as such. On the contrary, learned features commonly involve spikes of high amplitude which is consistent with the common reading of ictal activity. Figure \ref{fig:shap_compar} also shows some correlation of the shap values between channels F8-T8 and T8-P8. Correlation in the EEG signal may be detected by the model and learned as an ictal feature. However, it is not possible to distinguish whether the correlation is explicitly learned as a feature or if the same features are detected in both channels, resulting in similar shap values.

\section{Discussion}
In this work, the methodology was required to respond to several specifications. First, we aimed for an online seizure characterization method for potential wearable device applications. Second, the method needed to handle the high patient inter-variability in terms of EEG signals and seizure expression. Finally, we aspired to continue enlightening the black-box nature of DL models for seizure detection to meet the demand of clinical applications. We discuss the realization of the first two points in Section \ref{section:methodology}, and the last one in Section \ref{section:interpretability}.

\subsection{Evaluation of the methodology} \label{section:methodology}

We reduced the signal pre-processing steps to allow for an online detection with little delay. Raw EEGs from temporal electrodes are directly fed to the network, and the output probabilities were post-processed according to two different methods using Bayesian reasoning and a difference filter. To observe different modalities in terms of EEG processing, we trained three networks differing in the length of the first layer kernels. Results showed an antithetical behavior between the 3x131 and the 3x5 models in several steps of the methodology. First, at the segment-level, the 3x131 model showed a higher precision, while the 3x5 model showed a higher sensitivity even if both yielded similar F1-scores. This was also highlighted by different output probability distributions, where the 3x5 model was more shifted towards the positive class. In a situation with optimized hyper-parameters, these differences were blended. Yet, when fixing the hyper-parameters or observing the hyper-parameter space, the differences were still observable in predictions at the seizure-level. The main argument supporting this behavior is that the 3x131 model can explicitly extract low-frequency components in the first layer, which are often associated with resting potential EEGs and the interictal class. Depending on the application needed, one can exploit the opposing behavior of the two models to either reduce the false positive rate or to increase the sensitivity. 

The best performance overall was achieved by the 3x131 model after aggregating the segment-level predictions using the difference filter, achieving an F1-score of 0.873. If we look at the hyper-parameters space in Figure \ref{fig:heatmap_cv_diff} b), it seems to extend towards larger values of the filter size $M$. This extension is expected as the farther two samples are separated in time, in principle the stronger their difference will be and seizure events can be characterized more accurately. However, our method was constrained by the short recording time before seizure onset, and extending the interictal period would reduce the number of usable patients. This highlights the need for continuous time recordings for seizure events characterization, not only to study events leading to seizure onset but also to assess false-positive rates in the long term. The main motivation behind the difference filter method was to be able to detect changes in the pre-ictal parts of the signal. This would have meant that the network learned novel ictal features up to one minute before seizure onset and would be relevant for seizure onset prediction. Figure \ref{fig:post_process} shows that this method can detect seizure onsets mostly right after the true label. This suggests that drifts from an interictal EEG signal to a pre-ictal one is not detectable by training a classification model between samples 2 minutes before seizure onset and right after it. However, this post-processing method may be of particular relevance when training networks to differentiate pre-ictal and interictal segments. The difficulty of such a situation is that there is no evident time-separation between the two periods, which may reduce the training efficiency.

Figure \ref{fig:post_process} additionally shows that misdetected signals are generally consistent across all the three models suggesting that none of them was able to identify seizure events not identifiable by others. This is the common issue of patient inter-variability where the networks may not find the minimum in the loss hyperspace and sub-optimal more general solutions are preferred after training. In this sense, we observed that the main feature leading to ictal prediction was the EEG amplitude. Therefore, patients expressing seizure events without a significant amplitude variation may not be detected by any of the models. It would then be necessary to either increase the patient population for model training, or to apply transfer learning on a particular patient to adjust the model weights after a general training. Deeper models may also memorize such signals more efficiently, although up to date no evidence shows a better performance of deep networks over shallow ones on EEG classification \cite{schirrmeister_deep_2017}, \cite{roy_deep_2019}. 

\subsection{Decision interpretability} \label{section:interpretability}
We employed two methods to explore how the kernels of the first layer were contributing to the final decision and to highlight ictal features on the input EEG. Figure \ref{fig:max_input_filters} and Tables \ref{table:3x91_main_freq} and \ref{table:3x131_main_freq} show that some frequencies are strongly associated to the ictal class, while few ones are associated with the interictal class. In this latter case, the 8Hz frequency components are always associated with the interictal class, matching the common association of the alpha band with resting brain activity. This result shows that some features learned by the model are congruent with the current expertise on brain electrical activity. On the contrary, high-gamma frequencies and some beta band frequencies are commonly associated with the ictal class. It is expected that most features learned are necessary for the detection of ictal features, as they can easily be noticed to the naked-eye, while still learning few features such as low amplitude or resting brain activity as a counter-balancing information.

However, with this method it is not possible to infer that a frequency component in the maximized inputs is the solely cause leading to a specific classification. Shape, phase, and spatial combinations between channels may also be the key feature of a maximized input. For example, high frequencies in the maximized inputs may be responsible for keeping amplitude information across layers without losing frequency components in the input signals, but high frequency components by themselves may not carry explicit ictal information. To study whether high-gamma frequencies are needed for amplitude conservation or decision, one should also compare these results with normalized input models. If models trained with normalized inputs exhibit fewer high-gamma frequencies in the maximized inputs, this would be in favor of high-gamma frequencies being important to keep information relative to amplitude. Nevertheless, high amplitude patterns may be due to artifacts caused by movements during a seizure and not be specifically characteristic of the ictal phase. If one is interested in observing ictal features only, adding noisy EEG containing movements to the training set would facilitate the learning of ictal activity-specific features. Some other studies focusing on spectral bands for seizure classification showed that high-gamma frequencies were important features \cite{park_seizure_2011}. Other methods such as \cite{schirrmeister_deep_2017} have been developed to study the explicit link between the presence of a given frequency in a signal and its effect on the output decision.

\begin{figure}
    \centering
    \includegraphics[width = 0.9\textwidth]{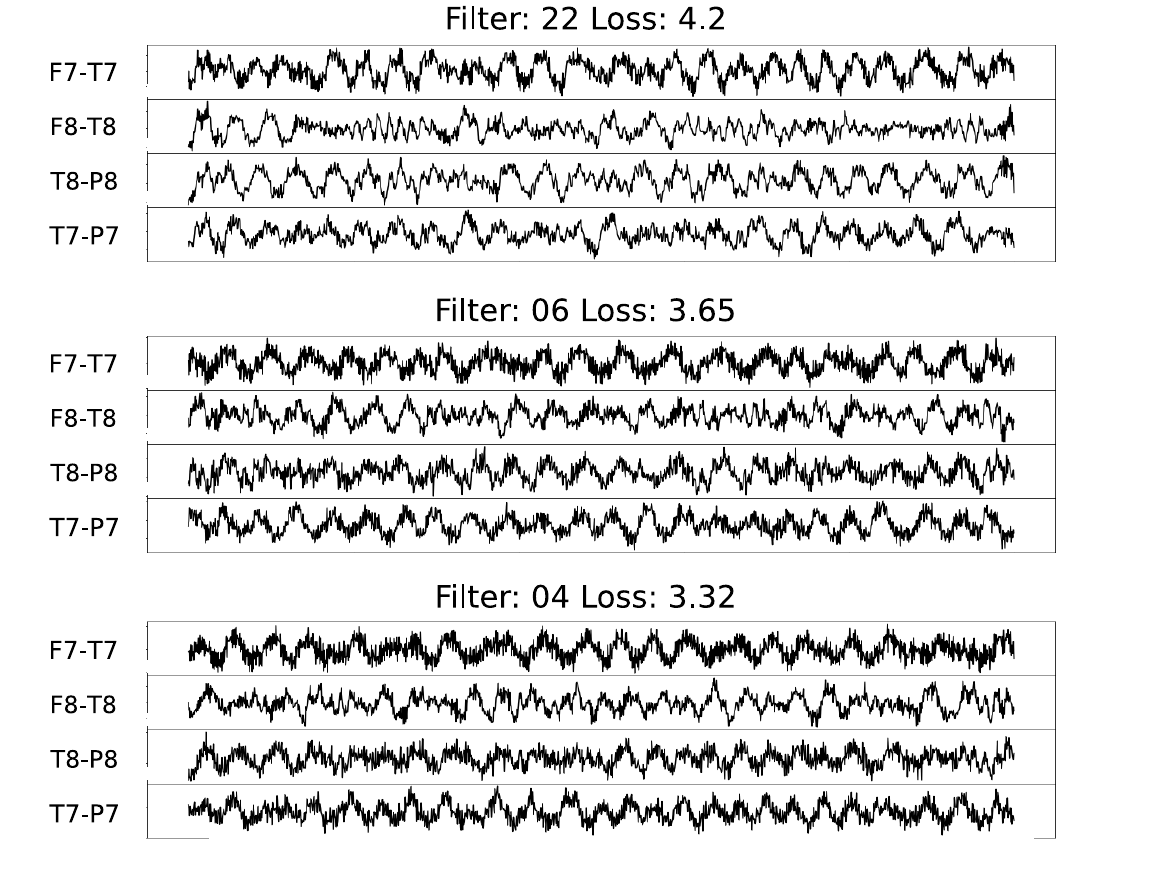}
    \caption{\textbf{Maximized inputs for kernels of the last convolutional block - 3x5 model.} The three maximizing inputs eliciting the strongest activation after the last convolutional block of the 3x5 model are represented. This shows that the 3x5 model learned to make implicit representation of low frequency components of the EEG signal.} 
    \label{fig:layer_6_maxinputs_3x5}
\end{figure}

Despite the 3x5 model having less explicit representation of the input data in the first layer, it performs as good as the other models in most cases. Indeed, Figure \ref{fig:max_input_filters} shows that all maximized inputs for the first layer kernels can hardly be interpreted and the power spectrum did not reveal any dominant frequency component. This goes in favor of amplitude being the most important feature learned by this network. This was verified in the maximized input experiment where the 3x5 models only had high-gamma frequency components that were shown to lead to ictal classification. However, Figure \ref{fig:layer_6_maxinputs_3x5} shows that after several convolutional blocks and pooling operations, the 3x5 model is able to construct an implicit representation of low-frequency components in the signal. Whether these components are used for predictions or if they are simple side-effects of the pooling operations remains to be elucidated. This model shows that even small kernel sizes can extract meaningful information for temporal binary classification over successive non-linear transformations and pooling of the input data. We should keep in mind that the 3x5 model has a 3x31 layer in the second block, and therefore it is able to explicitly extract frequencies as low as 9Hz. However, Table \ref{table:3x5_max_inputs_act6} shows that it can still learn frequency features as low as 4Hz. 

In Section \ref{section:methodology} we studied how the 3x5 models and 3x131 models were having similar performance while focusing more on sensitivity or specificity respectively. Maximized inputs bring some justification to this difference. As a kernel length of 5 samples along the time dimension can only detect frequencies above 51Hz, this model was focusing primarily on ictal features, and frequencies in the alpha band did not have explicit representation in the first layers of the model. On the contrary, both 3x131 and 3x91 models can detect frequencies as low as 2Hz and 3Hz respectively and could integrate interictal information in the first layers of the model. This may explain why the 3x5 model tends to have higher sensitivity and the output probability distribution is shifted towards the ictal class compared to the other two models.

\begin{table}
\begin{tabular}{lrrrrrr}
\toprule
\textbf{Filter idx} &        \textbf{F7-T7} &        \textbf{F8-T8} &        \textbf{T8-P8} &     \textbf{T7-P7} &      \textbf{pred} &      \textbf{loss} \\
\midrule
22 &    5.0 &    5.0 &    4.0 &    5.0 &  0.999394 &  4.174208 \\
6  &    5.0 &    5.0 &    4.0 &    5.0 &  0.999710 &  3.661172 \\
27 &    5.0 &    5.0 &    5.0 &    5.0 &  0.982527 &  3.294322 \\
\bottomrule
\end{tabular}
\caption{\textbf{Frequency components of top maximized inputs for last convolutional block of 3x5 models.}}
\label{table:3x5_max_inputs_act6}
\end{table}

On the other hand, the DeepLIFT visualization method gives informative visuals about the features learned by the models. Relating the highlighted features with the shape of the maximized inputs can further elucidate what has been learned by the model. Even if the maximized inputs suggest that amplitude may be an important feature, it is not sufficient nor necessary as some low amplitude patterns are detected as ictal in the DeepLIFT visualization, while some high amplitude patterns are not detected as such. High-gamma frequencies may also be responsible for detecting different shapes of spikes in the EEG signals, which seem to be the primary features highlighted on Figure \ref{fig:shap_compar}. An additional link regarding correlation between EEG channels and shap values can be suggested. Generalized seizures may result in synchronous brain activity which can be observed on EEG signals. As can be seen in Figure \ref{fig:max_input_filters}, some maximized inputs show a correlation of the shape of the signals between different channels. This phenomenon is also observed on the highlighted features of the DeepLIFT method as observed in Figure \ref{fig:shap_compar}. Since the first layer kernels span 3 channels at each pass on the input data, the identification of correlated patterns across channels is very likely to occur, especially with large kernel sizes in the time dimension. Further experiments may observe how the phase of the channels in the input data can influence the output probability, and hence grasp if a correlation across channels is an important feature learned by the model.

\section{Conclusions}
The goal of this study was to develop a DL-based methodology for online seizure event characterization able to handle inter-patient variability, and to explore some parameters of the model behavior from the interpretability point of view, including the problem of moving from a segment-level classification to a seizure-level classification. We demonstrated that the kernel size in the first layer is not significantly affecting the model performance, but a larger kernel size enables the study of the model behavior more thoroughly. We also provided insights on the features learned by the model by first observing the behavior of the first layer kernels and their maximized inputs and by highlighting the learned features back on the EEG input signal. Regarding the detection performance, our methodology was successfully able to generalize patient inter-variability for the majority of the population, and we found that the optimal time scale required for seizure-level classification is similar to that used by human experts when reading EEG signals. Moreover, the resulting model may be implemented in a wearable device with low energy requirements. Future developments should focus on the causality between important frequency components and the decision probability at the different internal states of the network and on handling classification of different sub-populations of seizures within a patient cohort to improve the generalization of the methodology.

\section*{Acknowledgement}

This work has been supported by the H2020 DeepHealth Project (GA No. 825111) and by the H2020 RECIPE project (GA No. 801137). The REPO$_2$MSE cohort was funded by the French Ministery of Health (PHRC national 2009) and sponsored by Hospices Civils de Lyon, and involved the following investigators: Philippe Ryvlin (Lyon, PI), Sylvain Rheims (Lyon, co-PI), Philippe Derambure (Lille), Edouard Hirsch (Strasbourg), Louis Maillard (Nancy), Francine Chassoux (Paris-St-Anne), Arnaud Biraben (Rennes), Cécile Marchal (Bordeaux), Luc Valton (Toulouse), Fabrice Bartolomei (Marseille), Jérôme Petit (La Teppe), Vincent Navarro (Paris-Salpétrière), Philippe Kahane (Grenoble), Bertrand De Toffol (Tours), Pierre Thomas (Nice).

\section*{Conflict of interest statement}
All authors declared no conflicts of interest


\bibliography{references}
\bibliographystyle{elsarticle-num}
\end{document}